\documentclass[parskip=full]{scrartcl}

\pdfoutput=1

\title{Research Trends and Applications of Data Augmentation Algorithms}

\author{%
	Joao Fonseca\(^{1*}\), Fernando Bacao\(^{1}\)
	\\
	\small{\(^{1}\)NOVA Information Management School, Universidade Nova de Lisboa}
	\\
	\small{*Corresponding Author}
	\\
	\\
	\small{Postal Address: NOVA Information Management School, Campus de
    Campolide, 1070--312 Lisboa, Portugal}
	\\
	\small{Telephone: +351 21 382 8610}
}

\usepackage{breakcites}
\usepackage{float}
\usepackage{graphicx}
\usepackage{geometry}
\usepackage{amsmath}
\usepackage{enumitem}
\usepackage[ruled,vlined]{algorithm2e}
\usepackage{booktabs}
\usepackage{pgfplotstable}
\pgfplotsset{compat=1.14}
\usepackage{longtable}
\usepackage{tabu}
\usepackage{hyperref}
\date{}

\usepackage{soul}

\definecolor{hypecol}{HTML}{0875b7}
\hypersetup{%
    colorlinks,
    linkcolor={hypecol},
    citecolor={hypecol},
    urlcolor={hypecol}
}

\begin{document}

\maketitle

\begin{abstract}

    In the Machine Learning research community, there is a consensus regarding
    the relationship between model complexity and the required amount of data
    and computation power. In real world applications, these computational
    requirements are not always available, motivating research on
    regularization methods. In addition, current and past research have shown
    that simpler classification algorithms can reach state-of-the-art
    performance on computer vision tasks given a robust method to artificially
    augment the training dataset. Because of this, data augmentation
    techniques became a popular research topic in recent years. However,
    existing data augmentation methods are generally less transferable than
    other regularization methods. In this paper we identify the main areas of
    application of data augmentation algorithms, the types of algorithms used,
    significant research trends, their progression over time and research gaps
    in data augmentation literature. To do this, the related literature was
    collected through the Scopus database. Its analysis was done following
    network science, text mining and exploratory analysis approaches. We
    expect readers to understand the potential of data augmentation, as well
    as identify future research directions and open questions within data
    augmentation research.

\end{abstract}


\section{Introduction}~\label{sec:introduction}

The performance of Machine Learning models is highly dependent on the quality
of the training dataset used~\cite{Fenza2021, Halevy2009}. Specifically, the
presence of imbalanced and/or small datasets, target labels incorrectly
assigned, outliers and high dimensional input spaces reduce the prospects of a
successful machine learning model implementation~\cite{Halevy2009,
Domingos2012, Salman2019}.  Even though the performance of any classifier is
affected by the size of its training dataset, deep learning models have a
particularly inconsistent performance over unseen datasets even when trained
with large datasets~\cite{Hu2020, Xie2021}.  Conversely, deep learning models
are capable of quickly adapting (and overfitting) to the training dataset,
including when it contains label and/or complete pixel noise~\cite{Xie2021,
Zhang2021}.  Although the performance of these models can be improved through
regularization methods, they are still incapable of correcting label noise in
the training dataset~\cite{Zhang2021}.

Regardless of the machine learning model used, when the training set contains
significant limitations (regarding overall quality and size), the model's
performance on unseen data is generally going to be affected. Specifically,
when the training data is not representative of the true population, or the
model is over-parametrized, it becomes particularly prone to
overfitting~\cite{Bartlett2021}. There are different strategies to reduce
overfitting, known as regularization methods~\cite{Shorten2019}. Identifying
the appropriate regularization methods varies according to the use
case~\cite{Chun2020}. While some methods can only be applied on specific
classifiers, data types or domains, others may be applied at the data level,
independently from the classification problem. For example, methods such as
dropout/dilution, batch normalization and transfer learning/domain adaptation
are mostly applied on neural network architectures.  Pruning is applied on
decision trees. Early stopping can be used on learners trained iteratively,
making it a broader method. 

\begin{figure}[H]
	\centering
	\includegraphics[width=.9\linewidth]{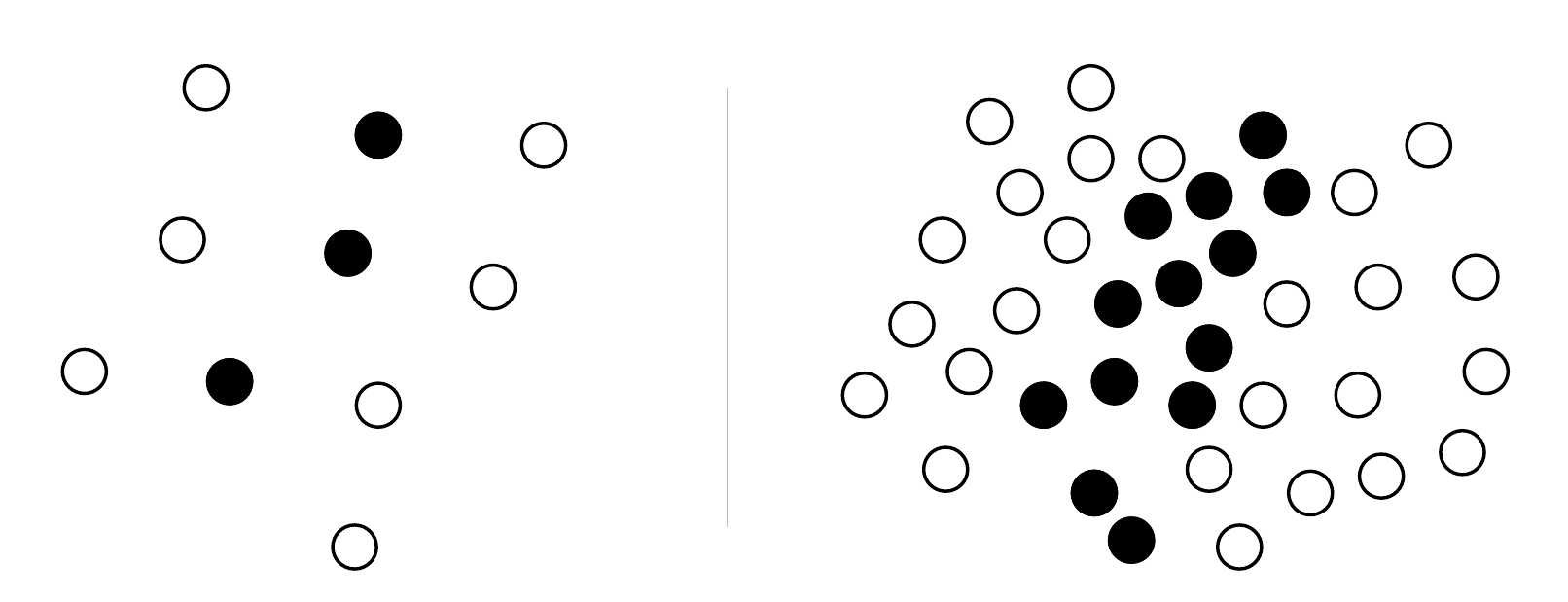}
    \caption{Example of data augmentation in a 2-dimensional binary
        classification setting. The left pane contains the original dataset,
        where the amount data is scarce and the gap among the two classes are
        wide, allowing for greater classification variability. The right pane
        contains the augmented dataset, where the gap among the two classes is
        narrower and the decision boundaries become easier to define.
    }~\label{fig:data_augmentation_example}
\end{figure}

Data augmentation techniques are used to increase the size (and hopefully the
diversity) of data in a training dataset through the production of artificial
observations~\cite{Van2001, Wong2016}. They are frequently used as
regularization techniques for various types of problems and classifiers, since
it is applied at the data level~\cite{Behpour2019}.
Figure~\ref{fig:data_augmentation_example} shows an example of data
augmentation, where the decision boundaries become clearer after the original
dataset is augmented. Data Augmentation methods can be divided into heuristic
and Deep Learning approaches~\cite{Shorten2019, Ratner2017}. Within these
approaches, they may be either domain specific or contextually independent.
For example, although both Synthetic Minority Oversampling Technique
(SMOTE)~\cite{Chawla2002} and Kernel Filters are heuristic approaches, SMOTE
may be used regardless of the context, while Kernel Filters are specific to
image data augmentation. The different types of Data Augmentation methods are
defined at a higher detail in Section~\ref{sec:data-augmentation}.

In 2011, Jürgen Schmidhuber's group showed that a MLP ensemble architecture
can achieve state-of-the-art performance on computer vision benchmarks given
strong enough data augmentation~\cite{Meier2011, Ciresan2011}. Although the
state-of-the-art improved since then, two recent papers developed by Google
Brain and Facebook research teams support Schmidhuber's group's findings.
Specifically, in~\cite{Tolstikhin2021, Touvron2021} the authors discuss two
similar MLP ensemble architectures, showing that the proposed model attains a
comparable performance to convolutional neural networks and attention-based
networks. Another recent study also discusses a related MLP architecture with
similar findings, while suggesting that the strong performance of computer
vision models may be attributable mainly to the inducive bias produced by the
patch embedding and the carefully-curated set of training
augmentations~\cite{Melaskyriazi2021}.

Research on data augmentation methods gained significant popularity in recent
years. As such, there were some efforts in the past to establish a taxonomy
and distinction of the different types of data augmentations
methods~\cite{Shorten2019}. To the best of our knowledge, there is no analysis
on data augmentation research as a whole, as well as domains of application
and future directions. In this paper we focus on current and past research
trends of data augmentation methods, its different applications and use cases.
This is done with an extensive analysis of the title, keywords and abstract of
a large set of literature related to data augmentation, collected through the
\href{https://www.scopus.com/}{Scopus} database using Natural Language
Processing and Network Science techniques. The analysis contains 3 phases. We
started by performing an exploratory data analysis to identify the most
significant publications, journals and conferences within the field of data
augmentation. Then, we analysed the articles' author keywords by constructing
a network and extracting and identifying communities of keywords.  Finally we
used a text mining approach to extract additional applications and methods
using the articles' abstracts, as well as validate the findings discussed with
the keyword analysis.

The rest of this paper is structured as follows:
Section~\ref{sec:data-augmentation} describes the main methods and approaches
used in data augmentation, Section~\ref{sec:methodology} describes the
procedures defined throughout the different analyses.
Section~\ref{sec:results_discussion} presents and discusses the findings drawn
from the analyses, as well as research gaps and open questions in data
augmentation research. Section~\ref{sec:conclusion} summarizes the main
findings discussed throughout the study.

\section{Data Augmentation Methods}~\label{sec:data-augmentation}
 
\begin{figure}[htb]
	\centering
	\includegraphics[width=\linewidth]{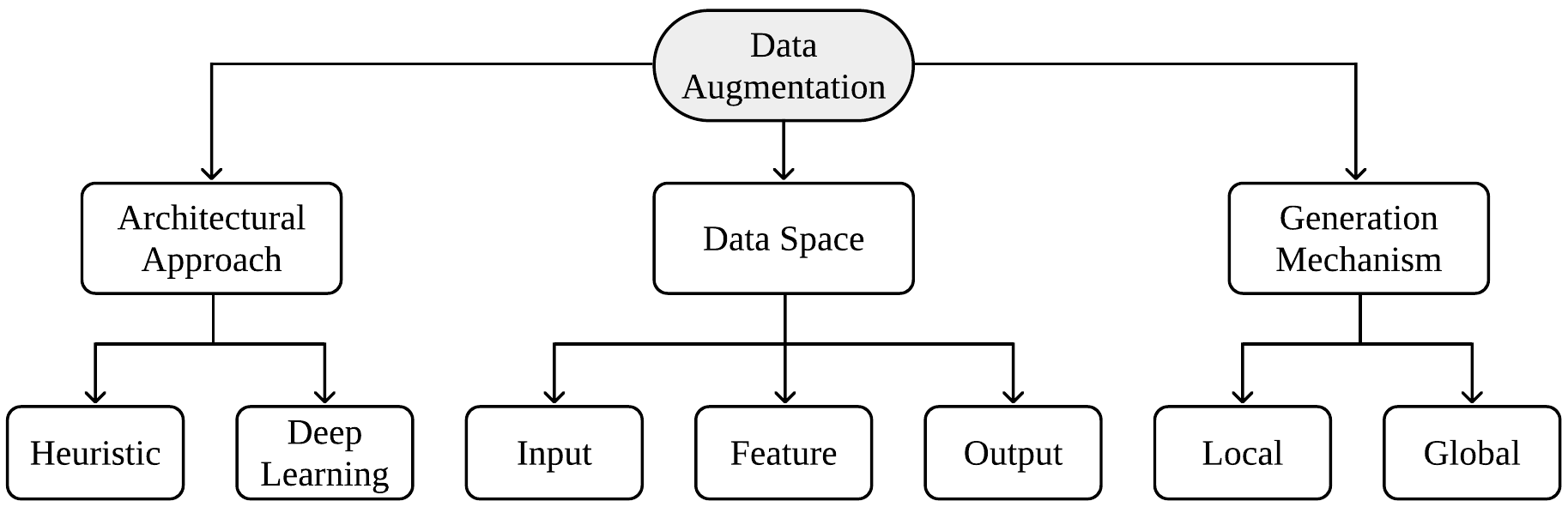}
    \caption{%
        Data Augmentation concept map.
    }~\label{fig:concept_map}
\end{figure}

Based on the literature found, a Data Augmentation method may be characterized
based on 3 criteria. The more common division is done between Heuristic and
Deep Learning approaches~\cite{Shorten2019}. Within these, several approaches
have been developed to produce artificial observations at the
input~\cite{Zhong2017}, feature~\cite{DeVries2017}, or output
space~\cite{Behpour2019}. Finally, we also distinguish them based on whether
their generation mechanism considers local (\textit{i.e.,} considers
partial/specific information within the dataset) or global (\textit{i.e.,}
it's based on the overall distribution/structure of the dataset) information
of the original dataset. Figure~\ref{fig:concept_map} depicts the concept map
with the different subdivisions of characteristics of data augmentation
methods. In this section, the analysis of the different types of data
augmentation will be based on their architectural approach. However, all the
methods mentioned may be divided using any of the definitions mentioned.

Heuristic methods use the information found in the input space to generate
new, relevant, non-duplicated observations by applying a predefined set of
rules, while incorporating a degree of randomness in the generation process.
Since data augmentation occurs in the input space, these are cost-effective
approaches to data augmentation. For this reason, heuristic methods are
simpler to implement and are particularly appealing for low dimensional
classification problems, especially when the computational power available is
limited. 

Some Deep Learning methods, on the other hand, attempt to map the original
input space into a lower-dimensional representation, known as feature
space~\cite{DeVries2017}. The generation of artificial observations occurs at
the feature space level, before being reconstructed to the original input
space. This is commonly done with Convolutional Neural Networks (CNN) and
auto-encoder architectures~\cite{Shorten2019}. Since data augmentation is
performed in the feature space, this type of approach is particularly useful
for high-dimensional data types and results in more plausible synthetic
observations~\cite{DeVries2017}. However, deep learning approaches require
more computational power than heuristic approaches and the resulting feature
space is difficult to interpret.

The difference in classification performance of the two perspectives is still
unclear. Wen et al.~\cite{Wen2020} evaluate the impact of data augmentation on
time series for various classification and forecasting tasks. Although they
found that both heuristic and deep learning approaches improved the results
over the various experiments, there was no direct comparison among the
different methods. Wong et al.~\cite{Wong2016} compared both input and feature
space data augmentation methods for image data classification performance over
the MNIST dataset. They found that input space augmentation can lead to better
classification performance if plausible transforms on the data are known.
However, in~\cite{DeVries2017} the authors discuss that the effectiveness of
each data augmentation method generally depends on the domain. The lack of
research on effective, domain-agnostic data augmentation methods appears to be
a current research gap.

\subsection{Heuristic Approaches}

Various heuristic approaches depend on the data type. For example, image data
augmentation may be done via translation, cropping or random
erasing~\cite{Zhong2017}, among others~\cite{Shorten2019}. However, these
techniques depend on the context and may not be applicable to other data types
such as time-series~\cite{Wen2020, Iwana2021} or tabular data. In this
subsection we will focus on domain-agnostic data augmentation methods.

Heuristic approaches may be applied at the input or feature space. The
appropriate method to be applied also depends on the machine learning goal.
Specifically, heuristic methods are commonly used to address classification
problems where the frequency of the different target classes vary
significantly, a problem known as Imbalanced Learning~\cite{Chawla2004}. In
this context, the dataset contains one or multiple rare classes, which become
more difficult to predict. This happens because during the learning phase,
classifiers are trained in order to maximize one or few performance metrics.
Although, a poor choice of performance metrics (\textit{i.e.}, a metric
insensitive to class imbalance, such as overall accuracy) might lead to a poor
estimation of the model's actual performance, since minority classes
contribute less to the learning phase and to the estimation of the performance
metric~\cite{Fonseca2021}. 

The problem of Imbalanced Learning is frequently addressed with oversampling
algorithms~\cite{Kaur2019}. Oversampling methods generate artificial
observations in order to balance class distributions using contextual
information based on the original dataset. These methods apply linear or
geometric interpolations between a random observation and one of its neighbors
to generate a new observation.

SMOTE~\cite{Chawla2002} is one of the most popular oversampling methods. It
generates an artificial observation along a line segment between a randomly
selected minority class observation and one of its nearest-neighbors. Since it
was first proposed, modifications at the neighbors parent observations
selection and data generation mechanisms of the original algorithm were
proposed. Borderline-SMOTE~\cite{Han2005} is an example of a modification of
SMOTE's data selection mechanism. Instead of selecting any random minority
class observation and one of its neighbors, the algorithm focuses in the
minority class observations closest to the decision boundary.
Geometric-SMOTE~\cite{Douzas2019} proposes a modification of the data
generation mechanism. Instead of generating data within a line segment, it
generates data within a hyper-spheroid between two parent observations.

Contrary to Deep Learning approaches, heuristic approaches can be applied at
the input space without the need of learning a feature space. This allows the
implementation of heuristic data augmentation with less computational power
and technical complexity. In addition, in contexts of limited data
availability (\textit{i.e.}, small datasets), deep learning approaches are not
appropriate since the amount of parameters to be tuned during the learning
phase often exceeds the number of observations in the dataset, making it
over-parametrized. The augmentation of small datasets using heuristic
approaches is explored in an Active Learning context
in~\cite{Fonseca2021active}. The authors found that significantly smaller
amounts of curated data using Active Learning, along with heuristic data
augmentation methods, achieved a classification performance comparable to
classifiers trained over the full dataset.

\subsection{Deep Learning Approaches}

Different deep learning approaches and architectures have been developed for
various domains. Deep learning data augmentation methods may be developed via
augmentation at the feature space (which involves learning a feature
space)~\cite{DeVries2017} or via a combination of a set number of observations
into a neural network in order to output a non-linear, non-geometric
combination of input observations~\cite{Wang2017}. Other domain-specific deep
learning methods also exist, such as style transferring techniques (specific
to image data)~\cite{Wang2017, Zhu2017}.

\begin{figure}[htb]
	\centering
	\includegraphics[width=.9\linewidth]{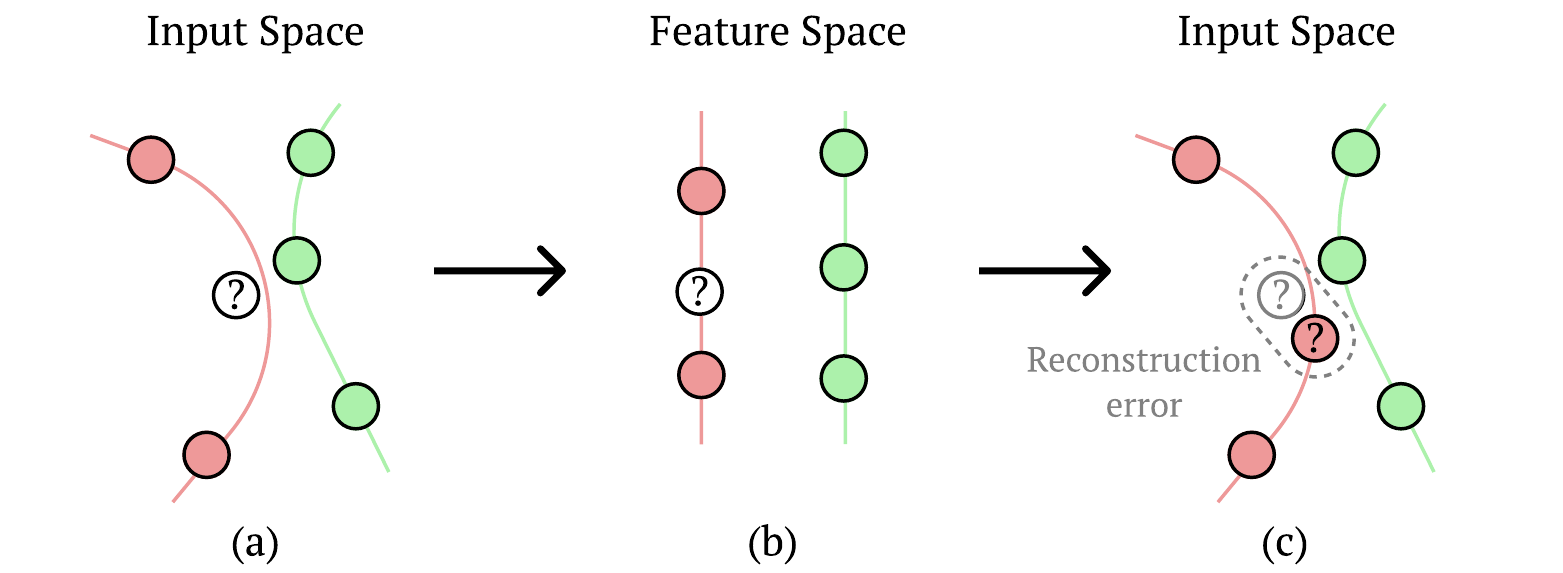}
    \caption{%
        Example of a sparse input space and its corresponding feature space.
        Learning a manifold feature space facilitates the generation of
        non-noisy artificial data. In the original input space (a) the
        unseen/artificial observation marked with ``?'' is closer to a green
        observation and to the learnt red manifold space, which may lead to a
        noisy observation. When projected or generated in the feature space
        (b), the observation will better match the nearest manifold measure
        and avoid noisy data in the input space (c). Adapted
        from~\cite{Antoniou2017}.
    }~\label{fig:input_vs_feature_space}
\end{figure}

Data augmentation at the feature space is especially useful when dealing with
high-dimensional datasets with complex and/or discontinuous distributions.
For example, many heuristic data augmentation techniques cannot be applied in
handwritten digits classification problems since they may change the true
label of the generated image and generate noisy data. In this situation,
performing data augmentation at the input space may introduce
noise~\cite{Chu2020}, since the data is also subjected to the curse of
dimensionality (see Figure~\ref{fig:input_vs_feature_space}a). This allows
transformations of known observations in a lower dimensional space to generate
new, non-noisy artificial observations projected in the input space, as shown
in Figures~\ref{fig:input_vs_feature_space}b
and~\ref{fig:input_vs_feature_space}c. 

The utilization of autoencoders~\cite{Kramer1991} is particularly useful to
perform feature space data augmentation~\cite{Shorten2019}. Autoencoders are
composed by an encoder and a decoder, which map the input space to and from
the feature space, respectively. The autoencoder is trained by minimizing the
difference between the original observation and the reconstructed observation
(see Figure~\ref{fig:input_vs_feature_space}c). Once the training phase is
completed, heuristic methods are applied in the feature
space~\cite{DeVries2017}.

Generative Adversarial Network (GAN)~\cite{Goodfellow2014} architectures is a
deep learning approach frequently used as a data augmentation method. It
involves a generator and a discriminator. Although many different
architectures have been proposed, the generator in the vanilla GAN algorithm
may be seen as a decoder, which is trained based on the gradients calculated
via the discriminator. The discriminator attempts to distinguish true and
generated (fake) observations in order to assess the quality of the data
produced by the generator. The problem is better formulated as a minimax
decision rule, where the generator attempts to fool the discriminator by
producing observations that are difficult to classify as generated.

When the size of the training dataset is not sufficiently large to employ deep
learning approaches and other related datasets or unlabeled datasets are
available, one technique that may also be used is transfer learning. In this
context, the data augmentation model is trained on a secondary model and is
later adjusted to the training dataset. This allows the usage of deep learning
models in few-shot learning environments~\cite{Antoniou2017}.

\section{Methodology}~\label{sec:methodology}

\begin{figure}[H]
	\centering
	\includegraphics[width=.85\linewidth]{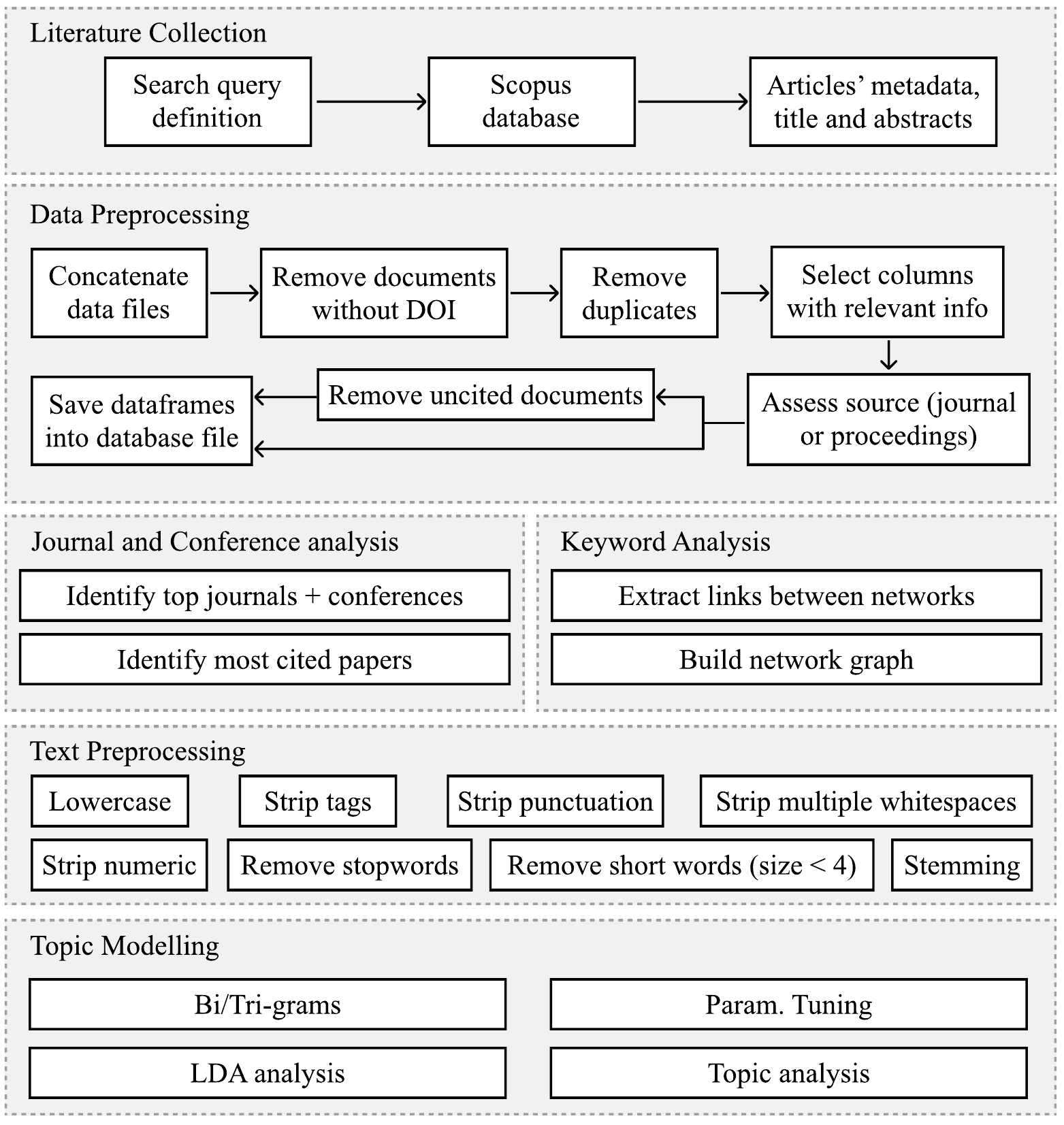}
    \caption{Diagram of the proposed literature analysis approach.
    }~\label{fig:slr_diagram}
\end{figure}

In this section we describe the procedures defined for the literature
collection, data preprocessing and literature analysis. The analysis of the
literature was developed with 3 different approaches. Throughout the
analyses, data preprocessing and hyperparameter tuning was developed
iteratively. The procedure adopted in this manuscript is shown in
Figure~\ref{fig:slr_diagram}.

The literature collection procedure is described in
Subsection~\ref{sec:lit_collection}. The data and text preprocessing is
described in Subsection~\ref{sec:data_preprocessing}. The exploratory data
analysis described in Subsection~\ref{sec:journal_and_conference_analysis} was
done to understand which manuscripts, journals and conferences are most
significant within the field of Data Augmentation. The manuscripts' keywords
were used to construct a network of keywords (described in
Subsection~\ref{sec:keyword_analysis}) and study the different communities of
keywords found in the network. The topic modelling and parameter tuning is
described in Subsection~\ref{sec:topic_modelling}. 

\subsection{Literature Collection}~\label{sec:lit_collection}

The focus of this literature analysis is to understand the different
algorithms, domains and/or tasks that employ data augmentation techniques.
Therefore, we search for documents containing the keyword ``data
augmentation'' in the search query. The results were then limited to
conference papers and journal articles written in English that were published
in the past 15 years.  Due to the large amount of results found, using solely
the \href{https://www.scopus.com/}{Scopus} database was found to be
sufficient. One of the goals during the search query design was to come up
with a simple and unbiased query. The resulting query is shown below:

\bigskip
\begin{verbatim}
KEY ("data augmentation") AND (LIMIT-TO (LANGUAGE, "English"))  
AND (LIMIT-TO (DOCTYPE, "cp") OR LIMIT-TO (DOCTYPE, "ar"))  
AND (
        LIMIT-TO (PUBYEAR, 2021) OR LIMIT-TO (PUBYEAR, 2020)  
    OR  LIMIT-TO (PUBYEAR, 2019) OR LIMIT-TO (PUBYEAR, 2018)  
    OR  LIMIT-TO (PUBYEAR, 2017) OR LIMIT-TO (PUBYEAR, 2016)  
    OR  LIMIT-TO (PUBYEAR, 2015) OR LIMIT-TO (PUBYEAR, 2014)  
    OR  LIMIT-TO (PUBYEAR, 2013) OR LIMIT-TO (PUBYEAR, 2012)  
    OR  LIMIT-TO (PUBYEAR, 2011) OR LIMIT-TO (PUBYEAR, 2010)  
    OR  LIMIT-TO (PUBYEAR, 2009) OR LIMIT-TO (PUBYEAR, 2008)  
    OR  LIMIT-TO (PUBYEAR, 2007) OR LIMIT-TO (PUBYEAR, 2006) 
)  
\end{verbatim}
\bigskip

The search query resulted in 4281 documents. The resulting data
selection/filtering pipeline is shown in
Figure~\ref{fig:data_filtering_pipeline}. Due to the limitations in the Scopus
data export (maximum 2000 documents per export), the data was split in four
different time periods and exported separately: 2006 until 2018, 2019, 2020
and 2021, which produced four CSV files.

\subsection{Data Preprocessing}~\label{sec:data_preprocessing}

\begin{figure}[H]
	\centering
    \includegraphics[width=.6\linewidth]{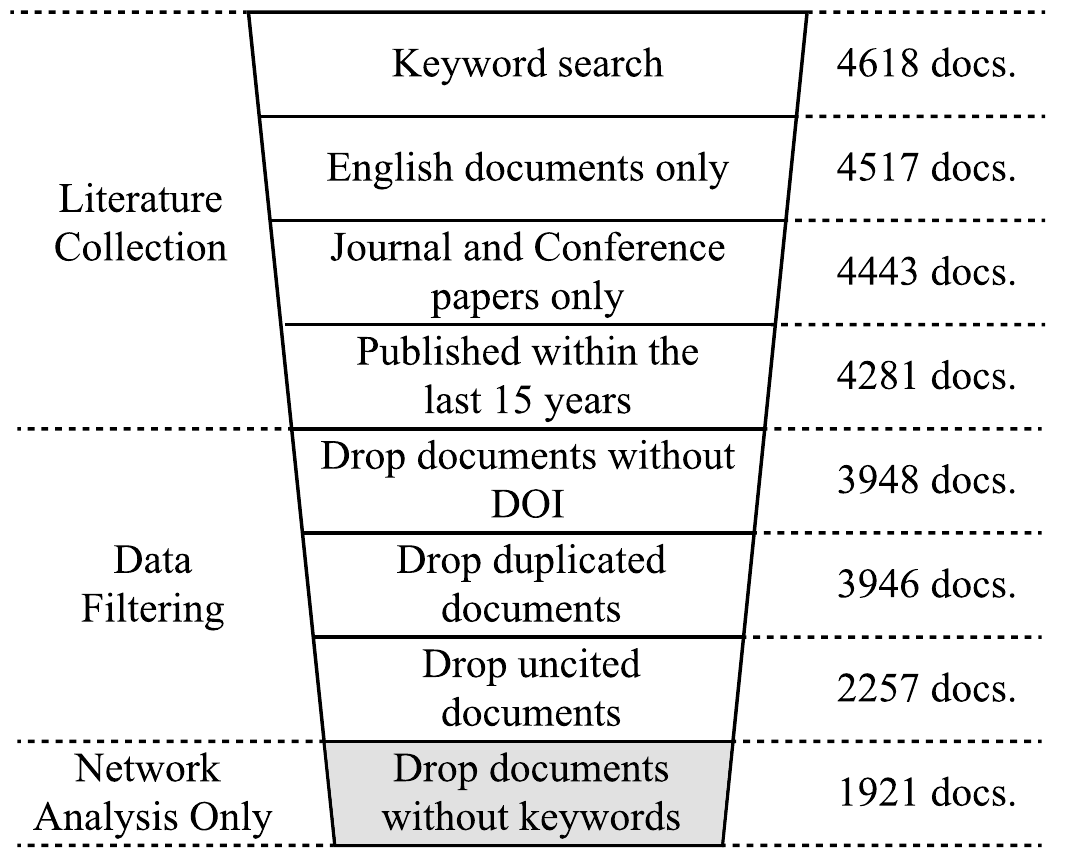}
    \caption{Data filtering pipeline.
    }~\label{fig:data_filtering_pipeline}
\end{figure}

The data preprocessing stage and amount of documents dropped is represented in
Figure~\ref{fig:data_filtering_pipeline}. The data was first concatenated into
a single data frame. During this process, we found that one of the exported
references had a corrupted line, which caused the loss of one additional
document.  Since the DOI can be used as a unique identifier for intellectual
property~\cite{Paskin1999}, references without a DOI were disregarded from
further analysis, while the ones with the same identifiers are removed
(\textit{i.e.}, only one of the repeating entries is kept).

This dataset was kept to perform the analysis described in
Subsection~\ref{sec:journal_and_conference_analysis}. However, further
preprocessing was done for the remaining parts of the literature analysis.
References without any citations were excluded for the keyword network and
topic modelling analyses. Finally, only the documents containing keywords in
Scopus' database were used to prepare the network analysis.

\subsection{Journal and Conference analysis}~\label{sec:journal_and_conference_analysis}

The exploratory analysis developed on the preprocessed dataset was targeted
towards the identification of the most significant works, journals and
conferences. We used the citation count as a proxy to understand the impact of
a specific manuscript within the research community.

The identification of the most significant conferences and journals is done by
sorting each type of publication according to the number of citations per
document. Conferences and journals with less than 10 papers published in the
area are not considered in this analysis. 

\subsection{Keyword Analysis}~\label{sec:keyword_analysis}

The analysis of keywords is expected to uncover general trends in data
augmentation research and its applications. The keyword ``data augmentation''
was removed since it would link with all other keywords. Keywords are
connected based on their co-occurrence in each research paper to form the
edges of the network.  It consists of an undirected graph whose weights are
based on the total citation count for the papers containing a given keyword
pair and is calculated as $\textrm{weight} = \log(\textrm{citations}) + 1$ to
avoid a potential bias caused by highly cited research articles. The size of
the nodes were determined with a logarithmic transformation of each
node's page rank.

Keyword combinations showing up in only one document are removed from further
analysis. The keyword network is then analysed using Python and the
communities were found using the greedy modularity maximization algorithm
proposed in~\cite{Clauset2004}. The results of the analysis and community
detection were ported to Gephi to produce the final visualizations.

\subsection{Topic Modelling}~\label{sec:topic_modelling}

The extraction of topics was done using the publication's abstracts. The words
were tokenized and all tags, special characters, punctuation, multiple white
spaces, numeric values, stop words and words with size smaller than 4 were
removed. Finally, we enriched the corpus by constructing bi-grams and
tri-grams.

We used a Latent Dirichlet Allocation (LDA) model~\cite{Pritchard2000} to
infer the topics present in our research domain. The tuning of the parameters
was done through experimentation and qualitative interpretation of the results
achieved. Additionally, the coherence score curve was also used as a reference for
parameter tuning and the choice of parameters, which are described in
Table~\ref{tab:hyperparameters}. 

\begin{table}[ht]
    \begin{center}
    \caption{Hyperparameters used.}~\label{tab:hyperparameters}
    \begin{tabular*}{.5\textwidth}{@{\extracolsep{\fill}}lllllll@{\extracolsep{\fill}}}
        \toprule
        Model   &   Hyperparameter  &   Value \\
        \midrule
        LDA     &   Num Topics      &   8     \\
                &   Chunk Size      &   2000  \\
                &   Passes          &   20    \\
                &   Alpha           &   0.1   \\
                &   ETA             &   auto  \\
        \bottomrule
    \end{tabular*}
    \end{center}
\end{table}

\subsection{Software Implementation}~\label{sec:software_implementation}

The analysis and modelling was developed using the Python programming
language, along with the
\href{https://scikit-learn.org/stable/}{Scikit-Learn}~\cite{Pedregosa2011},
\href{https://radimrehurek.com/gensim/}{Gensim}~\cite{Rehurek2010}, and
\href{https://networkx.org/}{Networkx}~\cite{Hagberg2008} libraries. The final
network analysis and visualization was done with
\href{https://gephi.org/}{Gephi}~\cite{Bastian2009}. All functions,
algorithms, analyses and results are provided in the
\href{https://github.com/joaopfonseca/ml-research}{GitHub repository of the
project}.

\section{Results \& Discussion}~\label{sec:results_discussion}

\begin{figure}[ht]
	\centering
    \includegraphics[width=\linewidth]{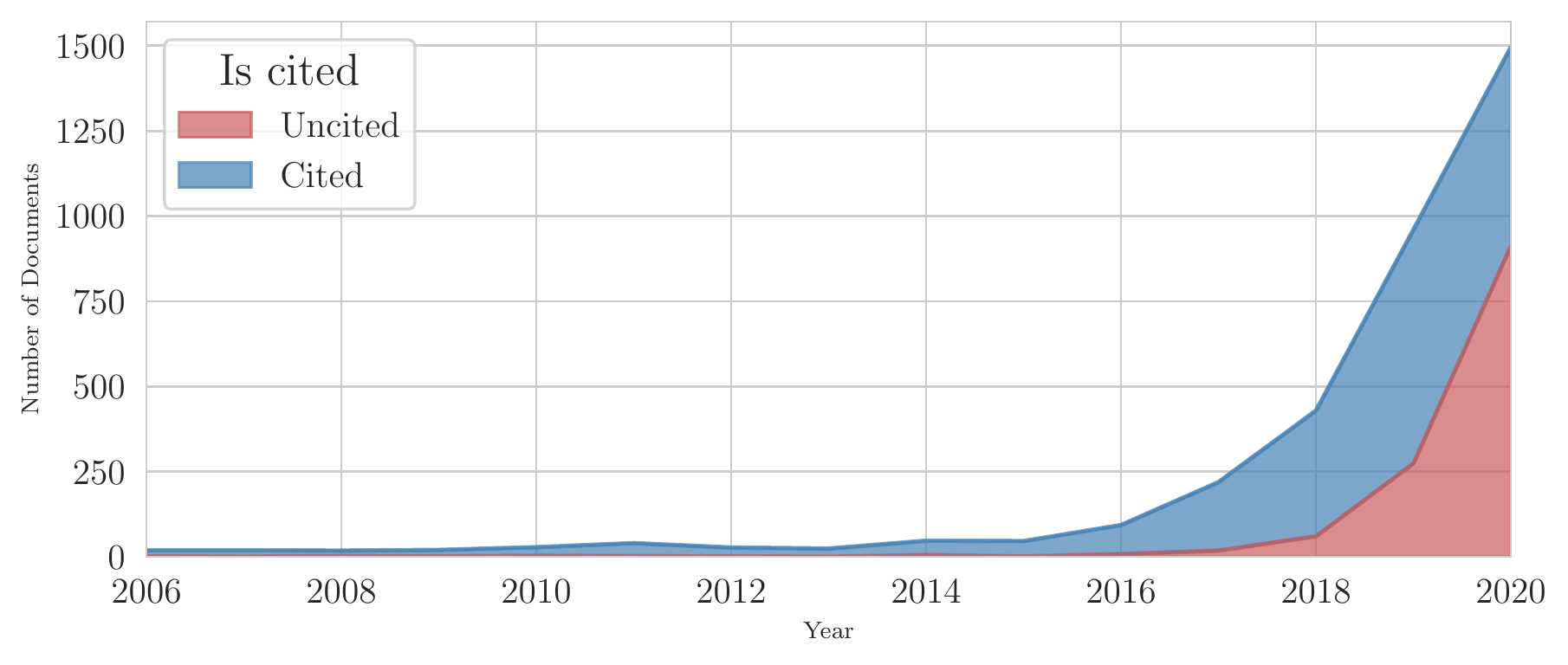}
    \caption{Annual number of publications containing the keyword ``data
        augmentation''.
    }~\label{fig:area_chart_cited_documents}
\end{figure}

The popularity of research in data generation has grown significantly in the
past 5 years, as shown in Figure~\ref{fig:area_chart_cited_documents}. Despite
the significant amount of uncited publications, out of the ones published in
2020, 39\% have already been used in other works. Although most of the
research developed before 2016 was used in other works, the amount of cited
research increased significantly after that period.

\subsection{Journal and Conference Analysis}

The initial exploration of the bibliometric data allows us to assess which
journals focused in data augmentation more intensely over the past years, as
shown in Table~\ref{tab:top_journals}. Most of the top journals belong to
technical fields, predominantly from Statistics, Remote Sensing, Medical
Imaging and other domains of applications such as agriculture. In addition,
all these journals have a high impact in their respective fields (based on
\href{https://www.scimagojr.com/}{Scimago Journal \& Country Rankings}).   

\begin{table}[ht]
    \begin{center}
    \caption{\label{tab:top_journals}
        Top journals focusing on data augmentation techniques, sorted by
        citations per document.
    }
    \begin{tabular*}{\textwidth}{@{\extracolsep{\fill}}lllllll@{\extracolsep{\fill}}}
        \toprule
        Source title & Publications & Citations & Average \\
        \midrule
        Journal of the American Statistical Association & 11 & 538 & 48.91 \\
        IEEE Geoscience and Remote Sensing Letters & 19 & 552 & 29.05 \\
        Neurocomputing & 35 & 808 & 23.09 \\
        Expert Systems with Applications & 14 & 283 & 20.21 \\
        Medical Image Analysis & 15 & 288 & 19.20 \\
        Neural Networks & 10 & 190 & 19.00 \\
        Journal of Computational and Graphical Statistics & 23 & 433 & 18.83 \\
        Computers and Electronics in Agriculture & 15 & 219 & 14.60 \\
        Biometrics & 13 & 163 & 12.54 \\
        IEEE Transactions on Medical Imaging & 10 & 123 & 12.30 \\
        \bottomrule
    \end{tabular*}
    \end{center}
\end{table}

Citation-wise, the publications coming from conference proceedings tend to
have a comparable impact in the research community, as shown in
Table~\ref{tab:top_conferences}. The most relevant conferences are positioned
in the computer science and information management fields. Research developed
in other areas of application, such as computer vision, speech recognition,
acoustic modelling, natural language processing and signal processing have
more activity in the form of conference proceedings publications. Conversely,
the domains most frequent in journal publications are not as active on
conference proceedings publications.

\begin{table}[ht]
    \begin{center}
    \caption{\label{tab:top_conferences}
        Top conferences focusing on data augmentation techniques, sorted by
        citations per document.
    }
    \begin{tabular*}{\textwidth}{@{\extracolsep{\fill}}lllllll@{\extracolsep{\fill}}}
        \toprule
        Source title & \# Pubs. & Cited & Avg \\
        \midrule
        Proceedings of the IEEE Computer Society Conference & 49 & 2111 & 43.08 \\
        \vspace{.2cm}on Computer Vision and Pattern Recognition &&& \\
        Lecture Notes in Computer Science (including subseries & 372 & 14946 & 40.18 \\
        Lecture Notes in Artificial Intelligence and Lecture Notes &&& \\ 
        \vspace{.2cm}in Bioinformatics) &&& \\

        \vspace{.2cm}Procedia Computer Science & 13 & 288 & 22.15 \\

        International Conference on Information and Knowledge & 10 & 180 & 18.00 \\
        \vspace{.2cm}Management, Proceedings &&& \\

        IEEE Computer Society Conference on Computer Vision & 23 & 314 & 13.65 \\
        \vspace{.2cm}and Pattern Recognition Workshops &&& \\

        ICASSP, IEEE International Conference on Acoustics, & 95 & 1153 & 12.14 \\
        \vspace{.2cm}Speech and Signal Processing - Proceedings &&& \\

        Proceedings - International Symposium on Biomedical & 30 & 346 & 11.53 \\
        \vspace{.2cm}Imaging &&& \\

        Proceedings of the International Conference on Document & 17 & 158 & 9.29 \\
        \vspace{.2cm}Analysis and Recognition, ICDAR &&& \\

        Proceedings of International Conference on Frontiers & 13 & 113 & 8.69 \\
        \vspace{.2cm}in Handwriting Recognition, ICFHR &&& \\

        2019 IEEE Automatic Speech Recognition and & 12 & 84 & 7.00 \\
        Understanding Workshop, ASRU 2019 - Proceedings &&& \\
        \bottomrule
    \end{tabular*}
    \end{center}
\end{table}

The papers with the highest citation count are listed in
Table~\ref{tab:top_papers}. We found that much of the research focused on
improving deep learning classification, segmentation or object detection
without a focus on a particular domain of application. Other papers centered
in the application of data augmentation methods for biomedical image
classification and segmentation, sound and speech recognition and remote
sensing.

\begin{table}[ht]
    \begin{center}
    \caption{\label{tab:top_papers}
        Top papers using data augmentation techniques, sorted by citation
        count.
    }
    \begin{tabular*}{\textwidth}{@{\extracolsep{\fill}}lllllll@{\extracolsep{\fill}}}
        \toprule
        Authors & Title & Year & Cited \\
        \midrule
        Ronneberger O., Fischer P., & U-net: Convolutional networks & 2015 & 13597 \\
        \vspace{.2cm}Brox T. & for biomedical image segmentation && \\

        Chatfield K., Simonyan K., & Return of the devil in the details: & 2014 & 1885 \\
        \vspace{.2cm}Vedaldi A., Zisserman A. & Delving deep into convolutional nets && \\

        Snyder D., Garcia-Romero D., & X-Vectors: Robust DNN Embeddings & 2018 & 636 \\
        \vspace{.2cm}Sell G., Povey D., Khudanpur S. & for Speaker Recognition && \\

        Shorten C., Khoshgoftaar T.M. & A survey on Image Data & 2019 & 590 \\
        \vspace{.2cm}                 & Augmentation for Deep Learning && \\

        Salamon J., Bello J.P. & Deep Convolutional Neural Networks & 2017 & 505 \\
                               & and Data Augmentation for \\
        \vspace{.2cm}          & Environmental Sound Classification \\

        Eitel A., Springenberg J.T., & Multimodal deep learning for robust & 2015 & 352 \\
        Spinello L., Riedmiller M., & RGB-D object recognition && \\
        \vspace{.2cm}Burgard W. &&& \\

        Ding J., Chen B., Liu H., & Convolutional Neural Network with & 2016 & 319 \\
        Huang M.                  & Data Augmentation for SAR Target && \\
        \vspace{.2cm}             & Recognition \\

        Wong S.C., Gatt A., & Understanding Data Augmentation & 2016 & 302 \\
        \vspace{.2cm}Stamatescu V., McDonnell M.D. & for Classification: When to Warp? && \\

        Frid-Adar M., Diamant I., & GAN-based synthetic medical image & 2018 & 296 \\
        Klang E., Amitai M., & augmentation for increased CNN && \\
        Goldberger J., Greenspan H. & performance in liver lesion && \\
        \vspace{.2cm}               & classification \\

        Bilen H., Vedaldi A. & Weakly Supervised Deep Detection & 2016 & 287 \\
        \vspace{.2cm}        & Networks \\

        \bottomrule
    \end{tabular*}
    \end{center}
\end{table}

\subsection{Keyword Analysis}

\begin{figure}[ht]
	\centering
    \includegraphics[width=\linewidth]{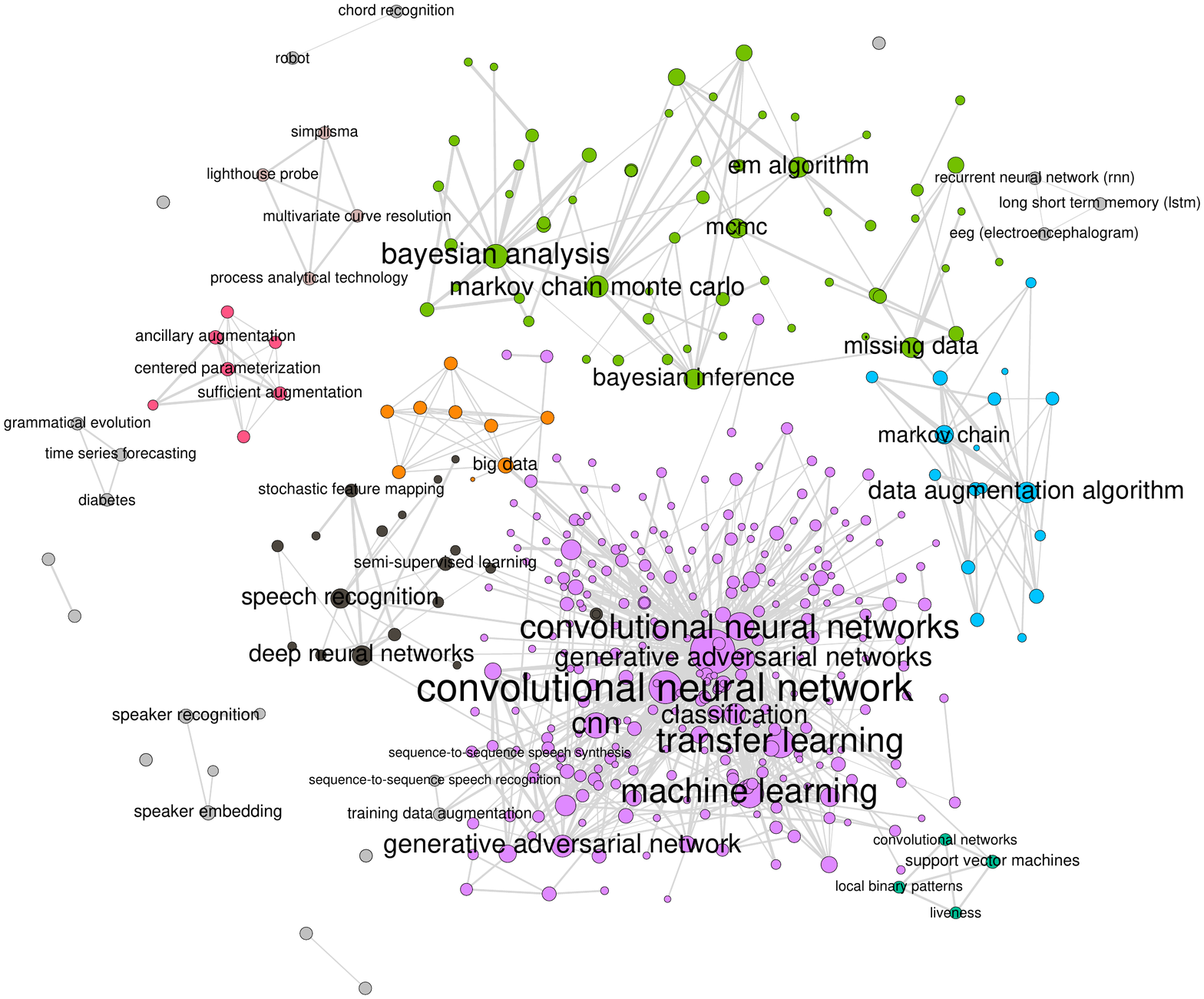}
    \caption{Keyword network.
    }~\label{fig:keyword_network}
\end{figure}

The keyword network shown in Figure~\ref{fig:keyword_network} revealed 8 main
communities of keywords, and 13 other small communities. The different
communities are distinguished by the type of algorithms used and/or the domain
of application. The main distinctive factor for the larger communities are the
types of generative models used, while the smaller communities are
distinguished according to the domain of application. The most significant
findings we found from this analysis are:

\begin{enumerate}
    \item The community marked with pink-colored nodes is characterized by the
        usage of neural network-based data augmentation methods in
        convolutional neural networks. The keyword ``deep learning'' is
        positioned as a central node (although not labelled in the figure to
        maintain readability). Other relevant keywords are related to
        machine/deep learning frameworks, deep learning classifiers and data
        augmentation algorithms, such as ``tensorflow'', ``keras'',
        ``convolutional neural network'' and ``generative adversarial
        networks''. Domain specific keywords are also present:
        \begin{itemize}
            \item Medical keywords located in this community cover a variety
                of applications. Relevant sub communities are [``hand
                writing'', ``parkinson's disease (pd)'', ``transfer
                learning''], [``breast cancer'', ``computer-aided
                detection''], [``melanoma'', ``skin cancer'', ``image
                processing'', ``googlenet''], [``chest x-ray'',
                ``computer-aided diagnosis'', ``tuberculosis'',
                ``segmentation''] and [``brain'', ``mri'', ``multiple
                sclerosis'']. 
            \item Remote sensing keywords are typically related to
                classification and object detection tasks. Relevant sub
                communities are [``object detection'', ``aerial image'',
                ``drone'', ``generative adversarial network'', ``semantic
                segmentation''], [``attributed scattering center (asc)'',
                ``synthetic aperture radar (sar)'', ``convolutional neural
                network (cnn)''], [``remote sensing'', ``road extraction'',
                ``transfer learning'', ``generative adversarial network''].
                Keywords such as ``hyperspectral imaging'' and ``weather
                classification'' are also scattered around the community.
            \item Facial recognition research is also represented in few sub
                communities: [``micro expression recognition'', ``small
                training data'', ``convolutional neural network (cnn)'', ``local
                binary pattern-three orthogonal planes (lbp-top)''] and
                [``training data augmentation'', ``sequence-to-sequence speech
                synthesis'', ``sequence-to-sequence speech recognition''].
            \item Fault detection studies also used data augmentation to deal
                with imbalanced datasets: [``fault diagnosis'', ``imbalanced
                data'', ``gan'']
            \item Data augmentation was also associated to regularization
                methods and feature extraction tasks, based on the presence of
                the sub communities [``overfitting'', ``dropout'' and ``cnn'']
                and [``feature extraction'', ``cnn'', ``svm''].
        \end{itemize}
    \item The community marked with blue-colored nodes is characterized by the
        usage of Markov Chain-based algorithms. The keywords ``markov chain'',
        ``data augmentation algorithm'' and ``monte carlo'' appear as central
        nodes. No application-specific sub-community was found.
    \item The community marked with green-colored nodes is characterized by
        the usage of Markov Chain and Bayesian-based algorithms. The keywords
        ``bayesian inference'', ``markov chain monte carlo'', ``mcmc'',
        ``bayesian analysis'', ``missing data'' and ``em algorithm''
        (expectation maximization algorithm). Application-specific keywords
        may be found sparsely distributed across the community, all of them
        related to biological applications. Specifically, the sub community
        [``ecological health'', ``stressor-response'', ``biological
        monitoring'', ``bayesian methods''] and the keyword ``camera
        trapping'' were found in this community. 
    \item The community marked with orange-colored nodes is characterized by
        keywords specific to big data and data warehousing applications. The
        network is composed of the keywords ``big data'', ``data lake'',
        ``olap'', ``map reduce'', ``cmm'', ``data warehouse'',
        ``augmentation'' and ``dm''.
    \item The remaining communities consist mostly of data augmentation
        methods applied to specific domains. Specifically, the usage of
        temporal-dynamic neural network architectures with ``eeg
        (electroencephalogram)'', music information retrieval applications
        (e.g., ``chord recognition''), speech/ speaker recognition and
        embedding, time series forecasting of diabetes and natural language
        processing and text classification.
\end{enumerate}

\subsection{Topic Analysis}

\begin{figure}[ht]
	\centering
    \includegraphics[width=\linewidth]{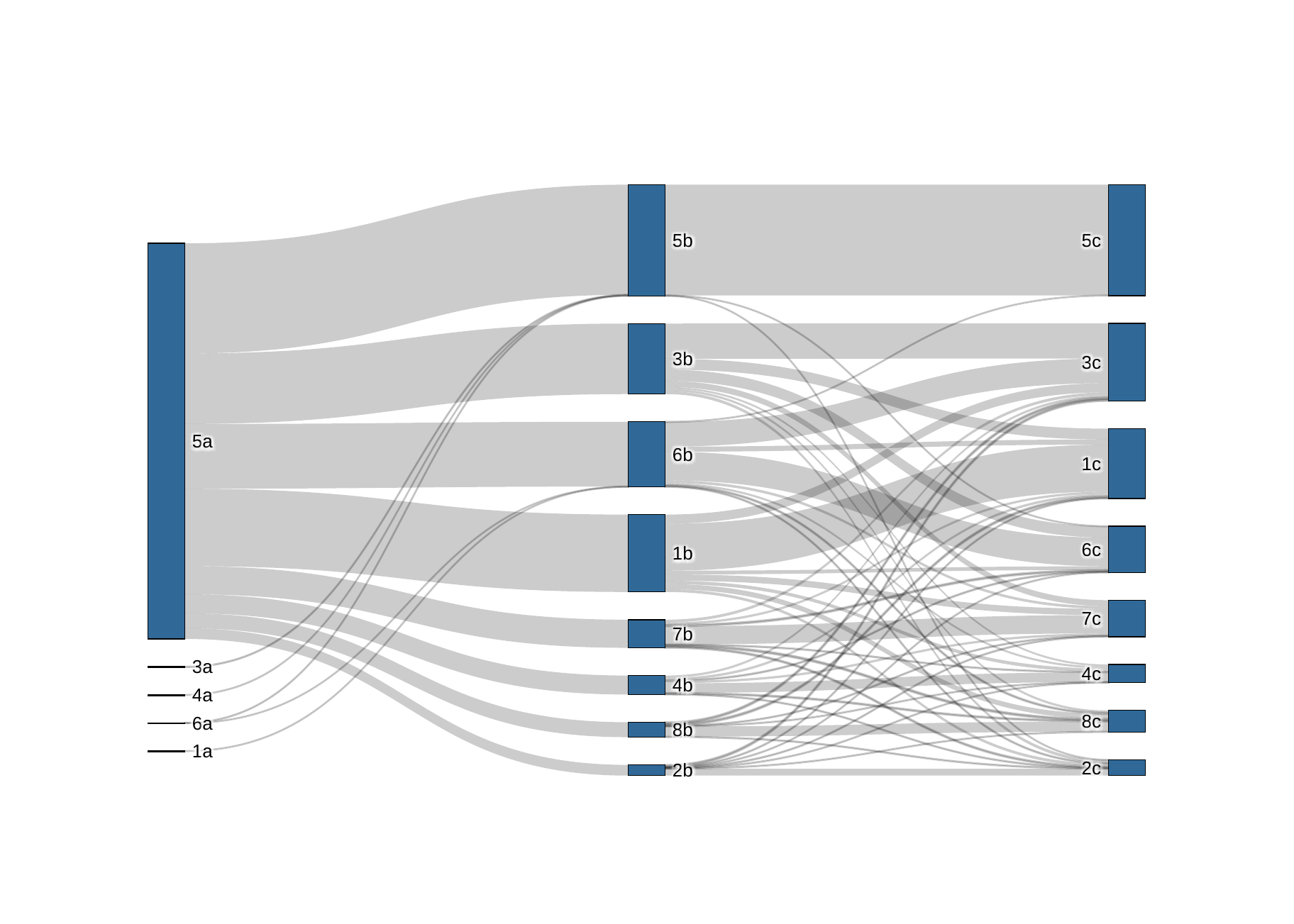}
    \caption{Distribution of documents over the different topics found. The
        left column represent the primary topics, the middle column represents
        the secondary topics and the right columns represents the tertiary
        topics.
    }~\label{fig:lda_topics_sankey}
\end{figure}

The LDA topic extraction resulted in 8 different topics, whose distribution of
topics is shown in Figure~\ref{fig:lda_topics_sankey}. The main topics within
which most articles were included is topic 5, which is defined by the main
theoretical keywords related to image data augmentation. Rather, the secondary
topic is more useful for this analysis. It is found based on the topic
likelihood of each document, excluding the dominant topic. Documents belonging
to the same group across primary, secondary and/or tertiary topics had a
likelihood of zero of belonging to any other topic.

The topics found in the bibliometric data are shown in
Table~\ref{tab:topic_analysis}. A few topics seem to overlap each other,
although they are generally distinguishable. The primary domains of
application of data augmentation methods differ for each different topic
identified:

\begin{enumerate}
    \item Documents in Topic 1 frequently use the word ``yolov'', which refers
        to the YOLOvX family of deep learning object detection
        models~\cite{Redmon2015}, where X refers to the version of the model
        used (the most recent version is 5). Another relevant keyword is
        ``style\_transfer'', which refers to a specific technique of data
        augmentation. 

        This topic has two primary domains of application. The keywords
        ``pest'' and ``coffe'' refer to data augmentation on agriculture
        research. The keywords ``biomed'', ``histolog'' and ``nodul'' refer to
        biomedical applications such as pulmonary nodule detection and
        histology image classification. Within these topics, a few
        domain-specific data augmentation algorithms were proposed. For
        example, in~\cite{Cicalese2020} the authors propose a style-transfer
        data augmentation method for histology image classification.  

    \item Documents in Topic 2 are primarily associated to the study of
        applications that include image data augmentation. The dominant
        keyword, ``hyperspectr\_imag'', refers to the application of data
        augmentation on hyperspectral images, commonly used in remote sensing
        and medicine. Other classification tasks include license plate
        detection (``licens\_plate''), inpainting (``inpaint''), background
        subtraction (``illumin\_chang'') and cloud shadow
        detection/segmentation (``shadow'').
    
    \item Documents in topic 3 refer to the application of data augmentation
        to deal with censored data (a condition in which the value of an
        observation is only partially known) and/or supervised tasks on data
        structured as graphs. Other domains of application involve chest
        x-rays classification (``cxr''), epidemiology (``risk\_factor'') and
        few audio/music information retrieval (``sourc\_separ'') articles.

    \item Documents in topic 4 refer to the application of data augmentation
        methods on object detection tasks. Specifically fire and smoke,
        pedestrians and crowd counting. Other applications within this topic
        are focused on speech recognition and angiography
        segmentation/classification.

    \item Documents in topic 5 are focused on image segmentation
        and classification methods where data augmentation algorithms are
        involved. It includes common keywords present in a large set of
        articles. These articles are mainly focused on the development of
        different convolutional neural network architectures (``cnn'') and
        neural network-based data augmentation methods.

    \item Documents in topic 6 are focused on Bayesian-based algorithms and
        Markov Chain algorithms. This topic includes data augmentation on
        regression tasks and misclassification detection.

    \item Documents in topic 7 covers the application of data augmentation
        into various domains. Specifically, music information retrieval
        (``music''), fish/marine organisms recognition, gender bias, speech
        recognition, random erasing 

    \item Documents in topic 8 contains remote sensing and biomedicine as the
        primary research domains. The keywords ``drone'' and ``aircraft''
        refer to the sources of data collected for remote sensing work,
        whereas ``pneumonia'' and ``chest\_rai\_imag'' refers to biomedicine
        research topics/image data.
\end{enumerate}

\begin{table}[ht]
    \begin{center}
    \caption{\label{tab:topic_analysis}
        Description of the main topics found in the literature.
    }
    \begin{tabular*}{\textwidth}{@{\extracolsep{\fill}}lllllll@{\extracolsep{\fill}}}
        \toprule
        Topic & Representative Paper & Papers & Words\\
        \midrule
        1 & GAN-based synthetic medical image & 440 & yolov, pest, style\_transfer, \\
          & augmentation for increased CNN && coffe, thermal, biomed, \\
          & performance in liver lesion && scene\_text, histolog, nodul, \\
        \vspace{.2cm}& classification && visibl \\
        
        2 & CVAE-GAN: Fine-Grained Image & 61 & hyperspectr\_imag, licens\_plate, \\
          & Generation through Asymmetric && command, inpaint, \\
          & Training && illumin\_chang, upper, restor, \\
        \vspace{.2cm}  &&& ann, foreign, shadow \\
        
        3 & A survey on Image Data & 401 & censor, markov\_chain, node, \\ 
          & Augmentation for Deep Learning && team, tree, cxr, risk\_factor, \\
        \vspace{.2cm}  &&&mass, largest, sourc\_separ\\
        
        4 & Return of the devil in the details: & 108 & smoke, pedestrian, transcrib, \\
          & Delving deep into convolutional nets && crowd, children\_speech, intent, \\
          &&&adult, auxiliari\_variabl, speech, \\
        \vspace{.2cm}  &&&angiographi \\
        
        5 & U-net: Convolutional networks for & 632 & imag, detect, gener, dataset, \\
          & biomedical image segmentation && classif, sampl, network, cnn, \\
        \vspace{.2cm}  &&&featur, augment\\
        
        6 & Deep Convolutional Neural Networks & 370 & tea, multivari, \\
          & and Data Augmentation for && markov\_chain\_mont\_carlo, \\
          & Environmental Sound Classification && bayesian, regress, misclassif, \\
        \vspace{.2cm}  &&& procedur, famili, illustr, mcmc \\
        
        7 & Weakly Supervised Deep Detection & 160 & music, fish, marin, gender, \\
          & Networks && vocal, random\_eras, low\_qualiti, \\
        \vspace{.2cm}  &&& crowd, prune, bengali \\
        
        8 & An Efficient Deep Learning Approach & 85 & drone, gait, aircraft, \\      
          & to Pneumonia Classification in && gestur\_recognit, pneumonia, \\
          & Healthcare && chest\_rai\_imag, covid, walk, \\
          &&& onset, hidden\_layer \\
        \bottomrule
    \end{tabular*}
    \end{center}
\end{table}

\begin{figure}[H]
	\centering
    \includegraphics[width=\linewidth]{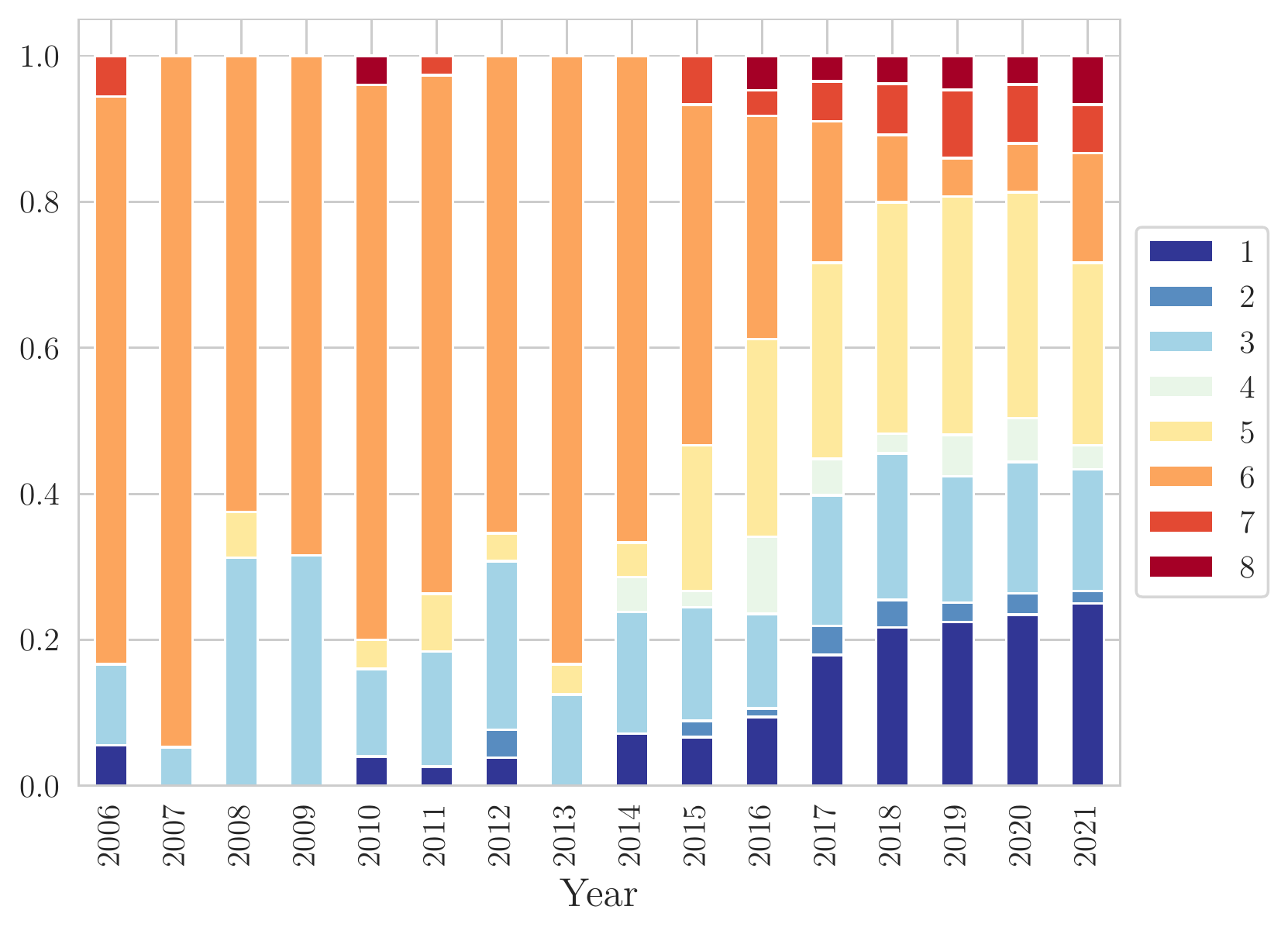}
    \caption{Topic frequency per year.
    }~\label{fig:topics_per_year}
\end{figure}

The per-year popularity of the different topics is shown in
Figure~\ref{fig:topics_per_year}. Since 2015, topic 5 gained more research
momentum, whereas topic 6 lost much of its relative popularity within the
field. In the past 5 years topics 8 and 3 have become steady research streams
while topic 1 saw a significant growth in popularity. 

\subsection{Research Gap Discussion}

Data augmentation mechanisms are often used as regularization methods for
deep learning classifiers. The study of data augmentation mechanisms in
ensembles of simple classifiers have achieved state-of-the-art performance not
only 10 years ago~\cite{Meier2011, Ciresan2011}, but also when compared to
modern deep learning architectures~\cite{Tolstikhin2021, Touvron2021,
Liu2021}. However, the implementation of different data augmentation methods
shows a promising path to improve the performance of simple classifiers
(and/or recent ensemble architectures) and requires further research.

A research application that was not frequently found in the literature was
small dataset augmentation. This is particularly useful for any complex
problem when the amount of labeled data available to use as training data is
scarce, which limits the usage of classification algorithms and especially
deep learning algorithms. In this context, techniques such as Active Learning
can be used to annotate a small amount of data, while maximizing the
classification performance~\cite{Su2020}. However, classifiers may not be
capable of generalizing with small training datasets and the ability to
reproduce and augment the labelled data available can further reduce
annotation cost and allow the usage of data intensive classifiers.

Another limitation found in the literature relates to the problem of
initialization on network-based data augmentation methods. The same data
augmentation algorithm trained with different initialization settings
(different random seed or training subset) may lead to different model
parameters and quality of the trained classifier.

The rapid development of data augmentation algorithms raises additional open
questions on how the data used and store for model training. Specifically, the
lower data storage and processing power available to the general public
(\textit{i.e.}, organizations and individuals) is a limitation for producing
state-of-the-art classifiers. Another problem arises from data privacy
concerns, since the usage of user data to train machine learning models
typically involve the storage of such information. However, if data
augmentation algorithms were continuously updated and capable of producing
reliable data on an as-needed basis, not only would storage requirements
decrease, but it would also become possible to work with fully artificial
data, without the need to store as much data. This would also facilitate the
sharing of datasets (in the form of an algorithm) without compromising
sensitive data.

\section{Conclusion}~\label{sec:conclusion}

Depending on the domain of application, data augmentation research differs in
the format of publication. On the one hand, domains like Statistics, Remote
Sensing and Medical Imaging seem more active on journal publications,
typically in journals with high impact factor. On the other hand, research
developed in the domains of computer vision, speech recognition, acoustic
modelling, natural language processing and signal processing seem to attribute
higher importance to conference papers. Many of the influential papers we found
were focused on deep learning methods for classification, segmentation, sound
and speech recognition and remote sensing.

We analysed the different communities of keywords formed using document
keywords, as well as topic analysis using a LDA analysis over the document's
abstracts. We found various distinctive areas of research, both regarding the
data augmentation methods used and the domain of application. We found that in
recent years research on augmentation methods using Bayesian-based algorithms,
as well as Markov Chain algorithms reduced its popularity, whereas data
augmentation methods based on neural networks and deep learning classifiers
have increased its popularity.

Data augmentation is most commonly applied/studied in the realm of computer
vision for tasks like image classification, segmentation, object detection,
inpainting and background subtraction tasks, even though it may be applied to
many other data structures. It is frequently used in studies within the
domains of biomedicine, agriculture, speech recognition, acoustic modelling,
remote sensing and computational creativity. It is also used alongside other
data preprocessing techniques, such as feature extraction and dimensionality
reduction.

Although data augmentation is a vibrant area of research, there are still
significant gaps to be addressed. Data augmentation methods are increasingly
used as regularization methods for deep learning. Although, recent research
shows that the same can be done for simpler classifier configurations in order
to achieve a classification performance comparable to that of state-of-the-art
deep learning, which require further confirmation, as well as the development
of less computational intensive data augmentation methods. Other less popular
topics, such as small data augmentation, appear to have a relevant practical
importance and require further research. In addition, other limitations of
data augmentation algorithms should be addressed. One problem commonly found
in the literature is the impact the weights initialization and training set
used have in the quality of the trained algorithm. In the future, using data
augmentation methods as a source of artificial datasets can address a variety
of concerns, such as data privacy, sharing and storage. Finally, exploring
data augmentation algorithms to complement or replace techniques such as
Active Learning may reduce the cost of data collection, although it is yet to
be explored.

\section*{Declarations}

\subsection*{Conflict of interest}

The authors declare that they have no conflict of interest or competing interest in
this work.

\subsection*{Authors' contributions}

Joao Fonseca collected all the research material and wrote the original text
of the paper. Fernando Bacao helped with the article preparation phase and
provided suggestions regarding the text, content and structure of the paper.

\subsection*{Availability of data and materials}

The data and the experiment's documentation is available at
\href{https://github.com/joaopfonseca/publications}{https://github.com/joaopfonseca/publications}.

\subsection*{Code availability}

The analyses and source code is available at
\href{https://github.com/joaopfonseca/publications}{https://github.com/joaopfonseca/publications}.

\bibliography{data-augmentation-trends}

\begin{thebibliography}{10}

\bibitem{Fenza2021}
G.~Fenza, M.~Gallo, V.~Loia, F.~Orciuoli, and E.~Herrera-Viedma, ``Data set
  quality in machine learning: Consistency measure based on group decision
  making,'' {\em Applied Soft Computing}, vol.~106, p.~107366, 2021.

\bibitem{Halevy2009}
A.~Halevy, P.~Norvig, and F.~Pereira, ``The unreasonable effectiveness of
  data,'' {\em IEEE Intelligent Systems}, vol.~24, no.~2, pp.~8--12, 2009.

\bibitem{Domingos2012}
P.~Domingos, ``A few useful things to know about machine learning,'' {\em
  Communications of the ACM}, vol.~55, no.~10, pp.~78--87, 2012.

\bibitem{Salman2019}
S.~Salman and X.~Liu, ``Overfitting mechanism and avoidance in deep neural
  networks,'' {\em arXiv preprint arXiv:1901.06566}, 2019.

\bibitem{Hu2020}
L.~Hu, C.~Robinson, and B.~Dilkina, ``Model generalization in deep learning
  applications for land cover mapping,'' {\em arXiv preprint arXiv:2008.10351},
  2020.

\bibitem{Xie2021}
Z.~Xie, F.~He, S.~Fu, I.~Sato, D.~Tao, and M.~Sugiyama, ``Artificial neural
  variability for deep learning: On overfitting, noise memorization, and
  catastrophic forgetting,'' {\em Neural computation}, vol.~33, no.~8,
  pp.~2163--2192, 2021.

\bibitem{Zhang2021}
C.~Zhang, S.~Bengio, M.~Hardt, B.~Recht, and O.~Vinyals, ``Understanding deep
  learning (still) requires rethinking generalization,'' {\em Communications of
  the ACM}, vol.~64, no.~3, pp.~107--115, 2021.

\bibitem{Bartlett2021}
P.~L. Bartlett, A.~Montanari, and A.~Rakhlin, ``Deep learning: a statistical
  viewpoint,'' {\em arXiv preprint arXiv:2103.09177}, 2021.

\bibitem{Shorten2019}
C.~Shorten and T.~M. Khoshgoftaar, ``A survey on image data augmentation for
  deep learning,'' {\em Journal of Big Data}, vol.~6, no.~1, pp.~1--48, 2019.

\bibitem{Chun2020}
S.~Chun, S.~J. Oh, S.~Yun, D.~Han, J.~Choe, and Y.~Yoo, ``An empirical
  evaluation on robustness and uncertainty of regularization methods,'' {\em
  arXiv preprint arXiv:2003.03879}, 2020.

\bibitem{Van2001}
D.~A. Van~Dyk and X.-L. Meng, ``The art of data augmentation,'' {\em Journal of
  Computational and Graphical Statistics}, vol.~10, no.~1, pp.~1--50, 2001.

\bibitem{Wong2016}
S.~C. Wong, A.~Gatt, V.~Stamatescu, and M.~D. McDonnell, ``Understanding data
  augmentation for classification: when to warp?,'' in {\em 2016 international
  conference on digital image computing: techniques and applications (DICTA)},
  pp.~1--6, IEEE, 2016.

\bibitem{Behpour2019}
S.~Behpour, K.~M. Kitani, and B.~D. Ziebart, ``Ada: Adversarial data
  augmentation for object detection,'' in {\em 2019 IEEE Winter Conference on
  Applications of Computer Vision (WACV)}, pp.~1243--1252, IEEE, 2019.

\bibitem{Ratner2017}
A.~J. Ratner, H.~R. Ehrenberg, Z.~Hussain, J.~Dunnmon, and C.~R{\'e},
  ``Learning to compose domain-specific transformations for data
  augmentation,'' {\em Advances in neural information processing systems},
  vol.~30, p.~3239, 2017.

\bibitem{Chawla2002}
N.~V. Chawla, K.~W. Bowyer, L.~O. Hall, and W.~P. Kegelmeyer, ``Smote:
  synthetic minority over-sampling technique,'' {\em Journal of artificial
  intelligence research}, vol.~16, pp.~321--357, 2002.

\bibitem{Meier2011}
U.~Meier, D.~C. Ciresan, L.~M. Gambardella, and J.~Schmidhuber, ``Better digit
  recognition with a committee of simple neural nets,'' in {\em 2011
  International Conference on Document Analysis and Recognition},
  pp.~1250--1254, IEEE, 2011.

\bibitem{Ciresan2011}
D.~C. Cire{\c{s}}an, U.~Meier, L.~M. Gambardella, and J.~Schmidhuber,
  ``Handwritten digit recognition with a committee of deep neural nets on
  gpus,'' {\em arXiv preprint arXiv:1103.4487}, 2011.

\bibitem{Tolstikhin2021}
I.~Tolstikhin, N.~Houlsby, A.~Kolesnikov, L.~Beyer, X.~Zhai, T.~Unterthiner,
  J.~Yung, A.~P. Steiner, D.~Keysers, J.~Uszkoreit, {\em et~al.}, ``Mlp-mixer:
  An all-mlp architecture for vision,'' in {\em Thirty-Fifth Conference on
  Neural Information Processing Systems}, 2021.

\bibitem{Touvron2021}
H.~Touvron, P.~Bojanowski, M.~Caron, M.~Cord, A.~El-Nouby, E.~Grave,
  G.~Izacard, A.~Joulin, G.~Synnaeve, J.~Verbeek, {\em et~al.}, ``Resmlp:
  Feedforward networks for image classification with data-efficient training,''
  {\em arXiv preprint arXiv:2105.03404}, 2021.

\bibitem{Melaskyriazi2021}
L.~Melas-Kyriazi, ``Do you even need attention? a stack of feed-forward layers
  does surprisingly well on imagenet,'' 2021.

\bibitem{Zhong2017}
Z.~Zhong, L.~Zheng, G.~Kang, S.~Li, and Y.~Yang, ``Random erasing data
  augmentation,'' in {\em Proceedings of the AAAI Conference on Artificial
  Intelligence}, vol.~34, pp.~13001--13008, 2020.

\bibitem{DeVries2017}
T.~DeVries and G.~W. Taylor, ``Dataset augmentation in feature space,'' 2017.

\bibitem{Wen2020}
Q.~Wen, L.~Sun, F.~Yang, X.~Song, J.~Gao, X.~Wang, and H.~Xu, ``Time series
  data augmentation for deep learning: A survey,'' {\em arXiv preprint
  arXiv:2002.12478}, 2020.

\bibitem{Iwana2021}
B.~K. Iwana and S.~Uchida, ``An empirical survey of data augmentation for time
  series classification with neural networks,'' {\em Plos one}, vol.~16, no.~7,
  p.~e0254841, 2021.

\bibitem{Chawla2004}
N.~V. Chawla, N.~Japkowicz, and A.~Kotcz, ``Special issue on learning from
  imbalanced data sets,'' {\em ACM SIGKDD explorations newsletter}, vol.~6,
  no.~1, pp.~1--6, 2004.

\bibitem{Fonseca2021}
J.~Fonseca, G.~Douzas, and F.~Bacao, ``Improving imbalanced land cover
  classification with k-means smote: Detecting and oversampling distinctive
  minority spectral signatures,'' {\em Information}, vol.~12, no.~7, p.~266,
  2021.

\bibitem{Kaur2019}
H.~Kaur, H.~S. Pannu, and A.~K. Malhi, ``A systematic review on imbalanced data
  challenges in machine learning: Applications and solutions,'' {\em ACM
  Computing Surveys (CSUR)}, vol.~52, no.~4, pp.~1--36, 2019.

\bibitem{Han2005}
H.~Han, W.-Y. Wang, and B.-H. Mao, ``Borderline-smote: a new over-sampling
  method in imbalanced data sets learning,'' pp.~878--887, 2005.

\bibitem{Douzas2019}
G.~Douzas and F.~Bacao, ``Geometric smote a geometrically enhanced drop-in
  replacement for smote,'' {\em Information sciences}, vol.~501, pp.~118--135,
  2019.

\bibitem{Fonseca2021active}
J.~Fonseca, G.~Douzas, and F.~Bacao, ``Increasing the effectiveness of active
  learning: Introducing artificial data generation in active learning for land
  use/land cover classification,'' {\em Remote Sensing}, vol.~13, no.~13,
  p.~2619, 2021.

\bibitem{Wang2017}
J.~Wang, L.~Perez, {\em et~al.}, ``The effectiveness of data augmentation in
  image classification using deep learning,'' {\em Convolutional Neural
  Networks Vis. Recognit}, vol.~11, pp.~1--8, 2017.

\bibitem{Zhu2017}
J.-Y. Zhu, T.~Park, P.~Isola, and A.~A. Efros, ``Unpaired image-to-image
  translation using cycle-consistent adversarial networks,'' in {\em
  Proceedings of the IEEE international conference on computer vision},
  pp.~2223--2232, 2017.

\bibitem{Chu2020}
P.~Chu, X.~Bian, S.~Liu, and H.~Ling, ``Feature space augmentation for
  long-tailed data,'' in {\em Computer Vision--ECCV 2020: 16th European
  Conference, Glasgow, UK, August 23--28, 2020, Proceedings, Part XXIX 16},
  pp.~694--710, Springer, 2020.

\bibitem{Antoniou2017}
A.~Antoniou, A.~Storkey, and H.~Edwards, ``Data augmentation generative
  adversarial networks,'' {\em arXiv preprint arXiv:1711.04340}, 2017.

\bibitem{Kramer1991}
M.~A. Kramer, ``Nonlinear principal component analysis using autoassociative
  neural networks,'' {\em AIChE journal}, vol.~37, no.~2, pp.~233--243, 1991.

\bibitem{Goodfellow2014}
I.~Goodfellow, J.~Pouget-Abadie, M.~Mirza, B.~Xu, D.~Warde-Farley, S.~Ozair,
  A.~Courville, and Y.~Bengio, ``Generative adversarial nets,'' {\em Advances
  in neural information processing systems}, vol.~27, 2014.

\bibitem{Paskin1999}
N.~Paskin, ``Toward unique identifiers,'' {\em Proceedings of the IEEE},
  vol.~87, no.~7, pp.~1208--1227, 1999.

\bibitem{Clauset2004}
A.~Clauset, M.~E.~J. Newman, and C.~Moore, ``Finding community structure in
  very large networks,'' {\em Phys. Rev. E}, vol.~70, p.~066111, Dec 2004.

\bibitem{Pritchard2000}
J.~K. Pritchard, M.~Stephens, and P.~Donnelly, ``{Inference of population
  structure using multilocus genotype data},'' {\em Genetics}, vol.~155, no.~2,
  pp.~945--959, 2000.

\bibitem{Pedregosa2011}
F.~Pedregosa, G.~Varoquaux, A.~Gramfort, V.~Michel, B.~Thirion, O.~Grisel,
  M.~Blondel, P.~Prettenhofer, R.~Weiss, V.~Dubourg, J.~Vanderplas, A.~Passos,
  D.~Cournapeau, M.~Brucher, M.~Perrot, and {\'{E}}.~Duchesnay,
  ``{Scikit-learn: Machine Learning in Python},'' {\em Journal of Machine
  Learning Research}, vol.~12, no.~Oct, pp.~2825--2830, 2011.

\bibitem{Rehurek2010}
R.~{\v R}eh{\r u}{\v r}ek and P.~Sojka, ``{Software Framework for Topic
  Modelling with Large Corpora},'' in {\em {Proceedings of the LREC 2010
  Workshop on New Challenges for NLP Frameworks}}, (Valletta, Malta),
  pp.~45--50, ELRA, May 2010.

\bibitem{Hagberg2008}
A.~A. Hagberg, D.~A. Schult, and P.~J. Swart, ``{Exploring Network Structure,
  Dynamics, and Function using NetworkX},'' in {\em Proceedings of the 7th
  Python in Science Conference} (G.~Varoquaux, T.~Vaught, and J.~Millman,
  eds.), (Pasadena, CA USA), pp.~11--15, 2008.

\bibitem{Bastian2009}
M.~Bastian, S.~Heymann, and M.~Jacomy, ``Gephi: an open source software for
  exploring and manipulating networks,'' in {\em Proceedings of the
  International AAAI Conference on Web and Social Media}, vol.~3, 2009.

\bibitem{Redmon2015}
J.~Redmon, S.~Divvala, R.~Girshick, and A.~Farhadi, ``You only look once:
  Unified, real-time object detection,'' in {\em Proceedings of the IEEE
  conference on computer vision and pattern recognition}, pp.~779--788, 2016.

\bibitem{Cicalese2020}
P.~A. Cicalese, A.~Mobiny, P.~Yuan, J.~Becker, C.~Mohan, and H.~Van~Nguyen,
  ``Stypath: style-transfer data augmentation for robust histology image
  classification,'' in {\em International Conference on Medical Image Computing
  and Computer-Assisted Intervention}, pp.~351--361, Springer, 2020.

\bibitem{Liu2021}
H.~Liu, Z.~Dai, D.~R. So, and Q.~V. Le, ``Pay attention to mlps,'' {\em arXiv
  preprint arXiv:2105.08050}, 2021.

\bibitem{Su2020}
T.~Su, S.~Zhang, and T.~Liu, ``{Multi-spectral image classification based on an
  object-based active learning approach},'' {\em Remote Sensing}, vol.~12,
  p.~504, feb 2020.

\end{thebibliography}
\bibliographystyle{ieeetr}
\end{document}